\documentclass{article}
\usepackage{spconf,amsmath,graphicx}
\usepackage{amssymb}
\usepackage{comment}
\usepackage{multicol}
\usepackage{multirow}
\usepackage{epsfig}
\usepackage{times}
\usepackage{url}
\ninept 
\title{Indian Masked Faces in the Wild Dataset}
\name{Shiksha Mishra$^{1}$ \quad Puspita Majumdar$^{1,2}$ \quad Richa Singh$^{1}$ \quad Mayank Vatsa$^{1}$}
  
  \address{$^{1}$ IIT Jodhpur, India \quad $^{2}$ IIIT-Delhi, India}
      
\begin{document}
%\ninept
%
\maketitle
\begin{abstract}
Due to the COVID-19 pandemic, wearing face masks has become a mandate in public places worldwide. Face masks occlude a significant portion of the facial region. Additionally, people wear different types of masks, from simple ones to ones with graphics and prints. These pose new challenges to face recognition algorithms. Researchers have recently proposed a few masked face datasets for designing algorithms to overcome the challenges of masked face recognition. However, existing datasets lack the cultural diversity and collection in the unrestricted settings. Country like India with attire diversity, people are not limited to wearing traditional masks but also clothing like a thin cotton printed towel (locally called as ``gamcha''), ``stoles'', and ``handkerchiefs'' to cover their faces. In this paper, we present a novel \textbf{Indian Masked Faces in the Wild (IMFW)} dataset which contains images with variations in pose, illumination, resolution, and the variety of masks worn by the subjects. We have also benchmarked the performance of existing face recognition models on the proposed IMFW dataset. Experimental results demonstrate the limitations of existing algorithms in presence of diverse conditions.
\end{abstract}
\section{Introduction}
\begin{figure}[t]
    \centering
    \includegraphics[scale=0.32]{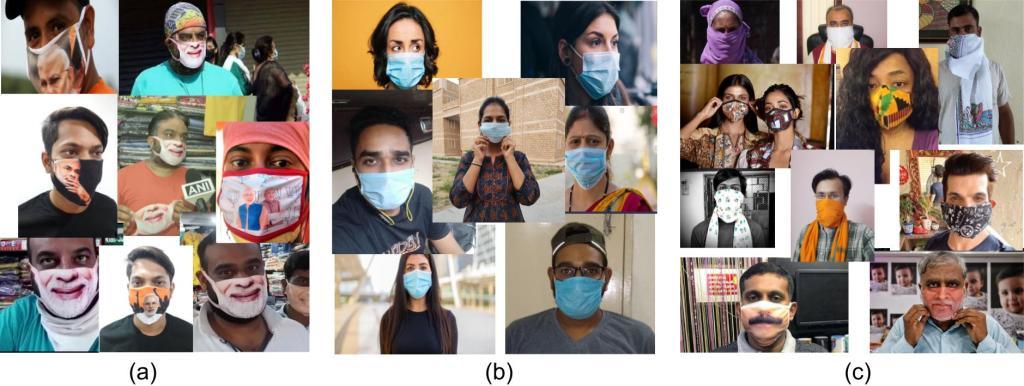}
    \caption{Illustration of various challenges in masked face recognition: (a) people wearing masks printed with facial images, (b) multiple people wearing same masks, and (c) different types of cloths are used as masks to cover the faces.} %Printed and graphical masks are in common use.}
    \label{fig:VisualAbstract}
\end{figure}

Face recognition algorithms have achieved tremendous success in handling different covariates such as low resolution, pose, expression, illumination \cite{hadsell2006dimensionality,bhattTIP2014, schroff2015facenet, deng2019arcface}. It is now used in various applications such as surveillance, access control, forensics, and e-payments and researchers have been enhancing the state-of-the-art performance on challenging tasks \cite{PAMI2017,fang2020generate,majumdar2020recognizing,zhang2019adacos}. However, the problem of recognizing faces under partial or heavy occlusion is still considered a challenging task \cite{zeng2020survey}. As the world is facing the pandemic of COVID-19, people worldwide are wearing masks as a protective measure. Wearing masks all the time has been completely normalized and also a mandatory step at workplaces. Masks occlude a fair amount of facial region, thereby making recognition difficult by face recognition algorithms. People wear different kinds of masks, from simple to the ones having many graphics that pose new challenges to automatic face recognition.  Fig. \ref{fig:VisualAbstract} shows some of these challenges in unconstrained masked face recognition, including the variation in the type of masks. 

Researchers have proposed several algorithms to overcome the challenges of occlusion in face recognition \cite{du2019nuclear, yuan2019face, li2020image, ge2020occluded}. However, very limited work has been done towards addressing the challenges of masked face recognition due to scarcity of masked face datasets. Ge et al. \cite{8099536} have proposed MAsked FAces dataset (MAFA) containing around 35,806 images of masked faces with varying pose and degree of occlusion. The dataset is created by downloading images from the Internet. MaskedFace-Net is a  recent dataset proposed by Cabani et al. \cite{cabani2020maskedface} that focuses on detecting masked faces as well as detecting whether the masks are worn correctly or not on the faces. Wang  et al. \cite{wang2020masked} have proposed three datasets, namely, MFDD, RMFRD, and SMFRD. MFDD is created for building robust detection algorithms. RMFRD is created by using web images, and SMFRD is prepared by simulating masks on the LFW \cite{huang2008labeled}, and Webface \cite{yi2014learning} datasets for recognition purposes. Naser et al.\cite{damer2020effect} have also proposed a dataset of 20 subjects to test the performance of face recognition algorithms. 

%Due to the ongoing research on masked face detection and recognition, some datasets have received the attention of the research community.

%Researchers have recently proposed some masked face datasets due to the ongoing research on masked face detection and recognition.

Existing datasets mainly include faces with Caucasian and North Asian demography with limited variations in the masks worn by the subjects. Some of them are collected in controlled settings. Further, none of the existing masked face datasets are built around the attire diversity, typically observed in Indian context. India is a diverse country, with people wearing different kinds of clothes such as stole or handkerchief as masks which pose new challenges to existing face recognition algorithms. Wearing printed masks with graphics is a common trend in India. This motivated us to propose a novel masked face dataset with Indian ethnicity, named as \textbf{Indian Masked Faces in the Wild (IMFW)} dataset. This dataset contains 200 subjects with both masked and non-masked images captured in an unconstrained environment. We believe that the proposed dataset will enable researchers to design sophisticated algorithms to overcome the challenges of masked face recognition in the Indian context. The following sections discuss the details of the proposed IMFW dataset, challenges in the IMFW dataset, and results of existing face recognition algorithms on the IMFW dataset.

 %In order to address this problem, we present a novel dataset, named as \textbf{Indian Masked Faces in the Wild (IMFW)} dataset for recognizing masked faces based on Indian ethnicity. This dataset contains 200 subjects with both masked and non-masked images captured in an unconstrained environment.

%The following sections discuss the challenges of masked face recognition, details of the proposed IMFW dataset, and results of existing face recognition algorithms on the IMFW dataset.

\section{Indian Masked Faces in the Wild}
\begin{figure}[]
    \centering
    \includegraphics[scale=0.3]{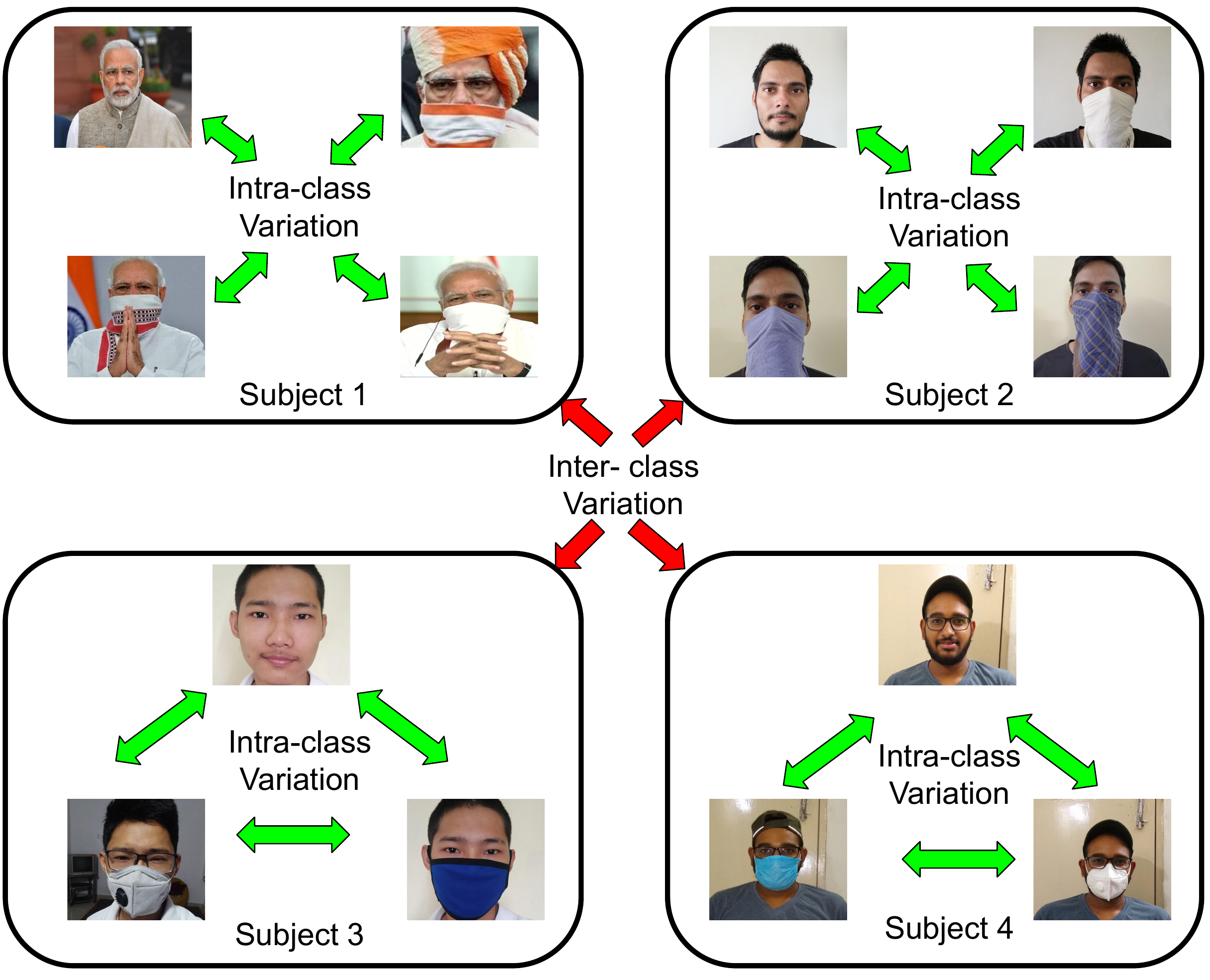}
    \caption{Summarizing the challenges of masked face recognition captured in the proposed IMFW dataset. All the subjects are representing diversity in demography and attire. Subject 1 and Subject 2 are showing the use of "gamcha" and "handkerchief" for covering faces.}
    \label{fig:InterIntra}
\end{figure}

We present the Indian Masked Faces in the Wild (IMFW) dataset of 200 subjects. The aim of this research is to overcome the challenges of unconstrained masked face recognition in the Indian context. Wearing ``gamchas'' and ``stoles'' as masks is a common trend in India. Variation in masks followed by the difference in skin tone variations pose new sets of challenges to face recognition as shown in Fig \ref{fig:InterIntra}. These variations are included in the proposed dataset to cover the diversity in India \footnote{The dataset and baseline models are available at: http://www.iab-rubric.org/resources/imfw.html}. The proposed IMFW dataset consists of three different sets with different number of subjects collected via different modes. For masked face recognition, face recognition algorithms are required to match the masked faces with the enrolled non-masked images. Therefore, for each subject, both masked and non-masked images are collected. Table \ref{Tab:Stats} summarizes the statistics of the proposed dataset. The following subsections discuss the details of the three sets in the proposed dataset.

\subsection{Set 1: Indian Celebrity}
\begin{figure*}[]
    \centering
    \includegraphics[scale=0.6]{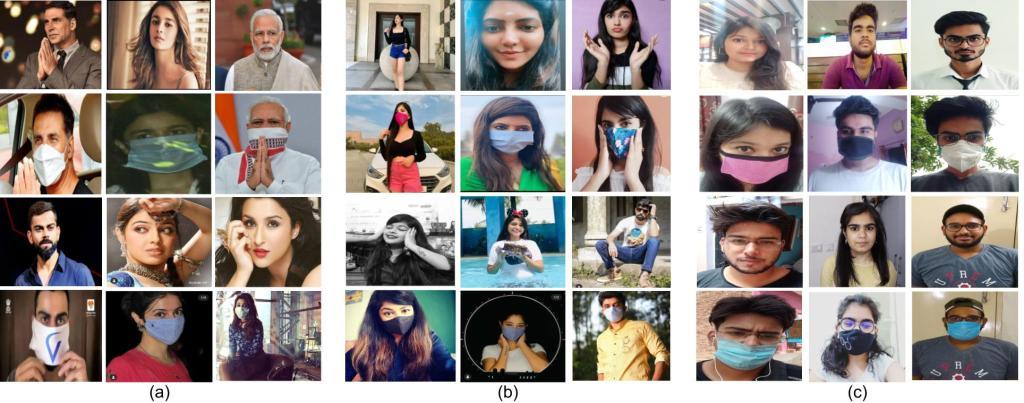}
    \caption{Sample images of the proposed IMFW dataset. (a) Set 1: Indian Celebrity, (b) Set 2: Instagram, and (c) Set 3: Indian Crowd. }
    \label{fig:DbCollage}
\end{figure*}

As shown in Fig. \ref{fig:DbCollage}(a), the Indian celebrity set of the proposed IMFW dataset contains 40 Indian celebrities with 435 images, including Bollywood actors/actresses, television stars, sports personalities, and politicians. These images are downloaded from the Internet. The majority of the downloaded images are taken in an unconstrained environment with varying poses and resolution. %Fig. \ref{fig:DbCollage}(a) shows some samples of the Indian Celebrity set.

\begin{table}[]
\centering
\renewcommand{\arraystretch}{1.2}
\caption{Statistics of the proposed IMFW dataset.}
\label{Tab:Stats}
\begin{tabular}{|l|c|c|c|}
\hline
\multicolumn{1}{|c|}{\multirow{2}{*}{\textbf{\begin{tabular}[c]{@{}c@{}}  \end{tabular}}}} &
  \multicolumn{2}{c|}{\textbf{No. of images}} &
  \multirow{2}{*}{\textbf{\begin{tabular}[c]{@{}c@{}}No. of\\  subjects\end{tabular}}} \\ \cline{2-3}
\multicolumn{1}{|c|}{} &
  \textbf{\begin{tabular}[c]{@{}c@{}}With\\  Mask\end{tabular}} &
  \textbf{\begin{tabular}[c]{@{}c@{}}Without \\ Mask\end{tabular}} &
   \\ \hline
\textbf{Set 1: Indian Celebrity}     & 214 & 221 & 40  \\ \hline
\textbf{Set 2: Instagram}            & 140 & 237 & 40  \\ \hline
\textbf{Set 3: Indian Crowd}         & 276 & 286 & 120 \\ \hline
\multicolumn{1}{|c|}{\textbf{Total}} & 630 & 744 & 200 \\ \hline
\end{tabular}
\end{table}

\begin{figure}[]
    \centering
    \includegraphics[scale=0.31]{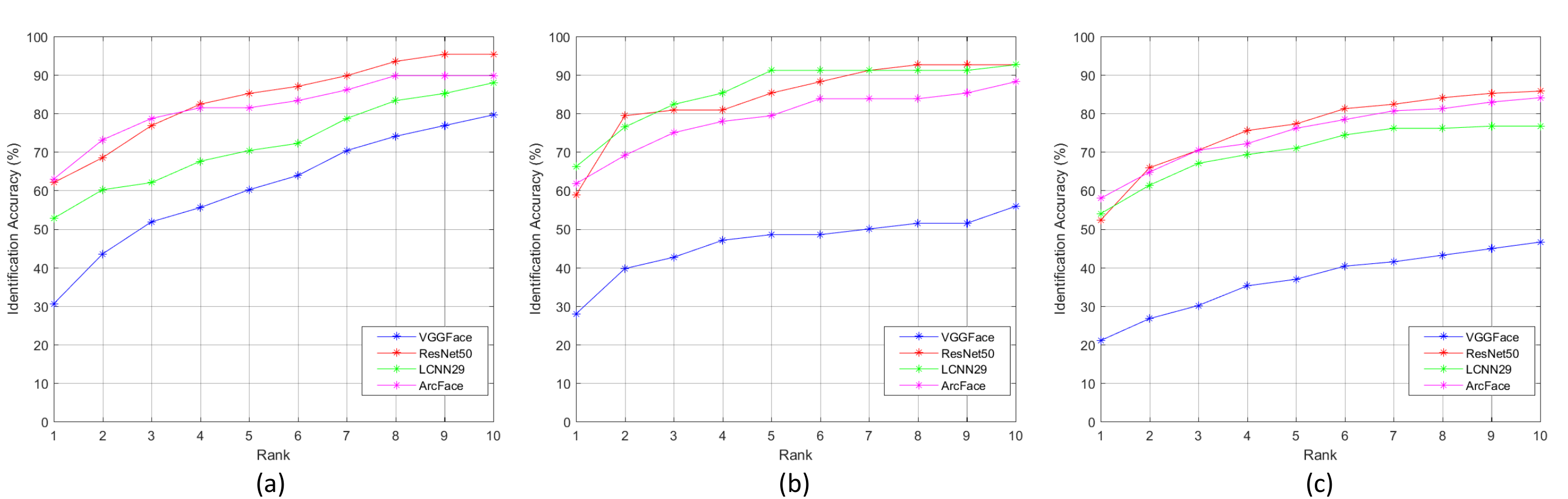}
    \caption{CMC curve of the four pre-trained deep face recognition models for masked face recognition on (a) Set 1 + Set 2, (b) Set 3, and (c) complete IMFW.}
     \label{fig:CMC}
     
\end{figure}

\begin{table*}[]
\centering
\renewcommand{\arraystretch}{1.2}
\caption{Identification accuracy (\%) of existing pre-trained deep face recognition models on the proposed IMFW dataset.}
\label{Tab:Pre-trained}
\begin{tabular}{|c|c|c|c||c|c|c||c|c|c|}
\hline
\multirow{2}{*}{\textbf{Models}} & \multicolumn{3}{c||}{\textbf{Set 1 + Set 2}}          & \multicolumn{3}{c||}{\textbf{Set 3}}                  & \multicolumn{3}{c|}{\textbf{Complete IMFW}}                   \\ \cline{2-10} 
                                     & \textbf{Rank 1} & \textbf{Rank 5} & \textbf{Rank 10} & \textbf{Rank 1} & \textbf{Rank 5} & \textbf{Rank 10} & \textbf{Rank 1} & \textbf{Rank 5} & \textbf{Rank 10} \\ \hline
\textbf{VGGFace}                     & 30.55           & 60.18           & 79.62            & 27.94           & 48.52           & 55.88            & 21.02           & 36.93           & 46.59            \\ \hline
\textbf{ResNet50}                    & 62.03           & 85.18           & 95.37            & 58.82           & 85.29           & 92.64            & 52.27           & 77.27           & 85.79            \\ \hline
\textbf{LCNN29}                      & 52.77           & 73.37           & 87.96            & 66.17           & 91.17           & 92.64            & 53.97           & 71.02           & 76.70            \\ \hline
\textbf{ArcFace}                     & 62.96           & 81.48           & 89.81            & 61.76           & 79.41           & 88.23            & 57.96           & 76.13           & 84.09            \\ \hline
\end{tabular}
\end{table*}

\begin{figure*}[]
    \centering
    \includegraphics[scale=0.6]{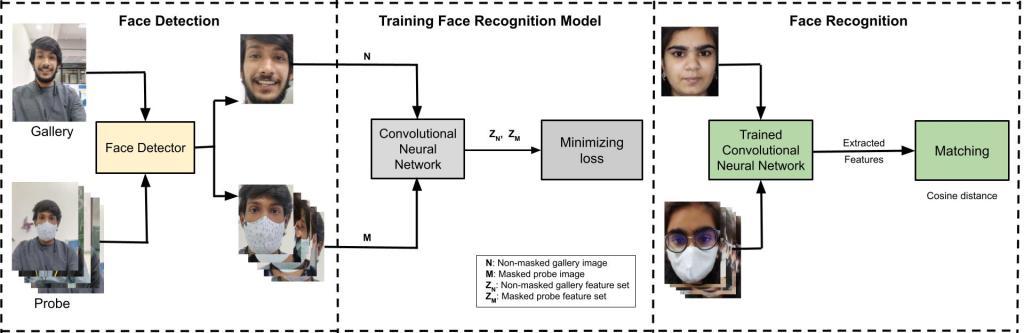}
    \caption{Pipeline of masked face recognition. The first block shows face detection using Tiny Face detector. Training of a face recognition model is shown in the second block. Third block shows face recognition using trained model.}
    \label{fig:Pipeline}
\end{figure*}

\subsection{Set 2: Instagram}
This set contains 377 images of 40 subjects downloaded from Instagram. We collected masked and non-masked images of Indian people with a public profile. Similar to Set 1, the majority of the images are taken in unconstrained settings, which includes variation in illumination conditions and backgrounds. Sample images of this set are shown in Fig. \ref{fig:DbCollage}(b).

\subsection{Set 3: Indian Crowd}
The images of this set are collected from the common people who volunteered to contribute to the dataset. Images are collected in both constrained and unconstrained environments. Images are captured using a mobile phone with a 48-megapixel rear camera and a 13-megapixel front camera. This set contains 120 subjects with 562 images. Apart from the unconstrained settings, variations in the type of masks used by the subjects are also considered during data collection. A wide variety of masks, including surgical, colored, N95, and printed, are used by the subjects. Some subjects have used stoles and handkerchiefs as masks during data collection to emulate the real-world scenarios. Fig. \ref{fig:DbCollage}(c) shows sample images of the Indian Crowd set.

\subsection{Challenges in the IMFW dataset}
%In the given scenario of the COVID-19 pandemic across the globe, it is highly recommended to wear masks as a protective measure. To motivate such step widely, there is a need to develop automatic face recognition algorithms that can recognize faces in the presence of masks. However, building such face recognition systems can be tricky because of the challenges posed by these masks, as mentioned below.

The proposed IMFW dataset is collected under unconstrained settings with large variations in pose, background, illumination, resolution, and the type of masks worn by the subjects. Images are captured in different lighting conditions with varying backgrounds. Additionally, the images have large pose ($-90^0$ to $+90^0$) variations. Variation in resolution (low to high) of the images is also considered during data collection. Multiple constraints present in the dataset, along with the diversity in the subjects with respect to demography, attire pose challenges to face recognition algorithms. Further, the masks worn by the subjects increase the difficulty of recognizing faces by automated systems. The challenges of face recognition in the presence of masks are:

%Our dataset was collected under unconstrained settings. These setting include variation in pose, background, illumination, resolution, and masks worn by people. The pose of a person may vary from +90 degree to -90 degree. Also, there is no definite background settings for taking the images of subject, The background may vary from very low contrast to very high contrast from the foreground, and may contain people in it also. The resolution of images in the dataset can vary from low to high. The masks worn by people ranges from very simple plane masks like surgical mask, or can have certain graphics in it. Not only that, there are also images where people are covering their faces with other clothing like, "gamcha", "handkerchief", or "stole". These variations present challenges for face recognition.

\noindent \textbf{Occlusion}: Wearing masks occlude a major section of the face leaving only eyes and forehead visible. Sometimes people wear eyeglasses/sunglasses that further occludes their faces. Such occlusion leaves the recognition task to be done by the forehead region only. In addition to this, unconstrained settings like the variation in pose, illumination, and resolution further exaggerate the challenges of automatic face recognition.
    
\noindent \textbf{Inter-class and intra-class variations}: As masks have become an integral part of our wardrobe, people accessorize themselves with a variety of masks on a regular basis. This variety of masks used by a subject results in intra-class variation among the samples of the same subject. Also, some masks like surgical masks are very common across the globe. These masks worn by different subjects decrease the inter-class separability. These issues lead to additional challenges in masked face recognition. 
    
\noindent \textbf{Masks with printed faces}: Recently, printed masks with face images have attracted the attention of some sections of the population. Some of these masks either contain the image of the lower half of a face or the complete face image. Such variation in masks further complicates the task of face recognition. Wearing such masks may hamper the performance of face recognition models.

%Additionally, these masks can be used as a disguised tool to impersonate someone's identity by wearing the mask with the target person's face printed on it. Wearing such masks may unintentionally fool the face recognition systems as well.

\section{Experimental Results}
The performance of existing deep face recognition models is evaluated on the proposed IMFW dataset. For this purpose, two different experiments are performed. The first experiment is performed to evaluate the performance of four pre-trained deep face recognition models namely, VGGFace \cite{parkhi2015deep}, ResNet50 (trained on the VGGFace2 dataset) \cite{cao2018vggface2}, LightCNN29 \cite{wu2018light}, and ArcFace \cite{deng2019arcface} on the proposed IMFW dataset. The second experiment evaluates the performance of existing loss functions, Contrastive loss \cite{hadsell2006dimensionality} and triplet loss \cite{schroff2015facenet} for masked face recognition.

\noindent \textbf{Protocol:} Experiments are performed by dividing the IMFW dataset into training and testing sets with non-overlapping subjects. Each of the three sets (Indian Celebrity, Instagram, Indian Crowd) is split into training and testing partitions with 70\% subjects in the training set and 30\% subjects in the testing set. Further, the training and testing sets are divided into gallery and probe. Experiments are performed to emulate the real-world scenario of matching masked face images with non-masked enrolled images. Therefore the gallery contains non-masked images, and the probe contains masked face images of each subject. The gallery contains a single image per subject, while the probe contains multiple images per subject. 

\noindent \textbf{Implementation Details:} LightCNN29 is used as the base network for model training using contrastive loss and triplet loss. Initial layers of the models are frozen, and the last ten layers are trained by minimizing the existing loss functions. Models are trained for 50 epochs with a learning rate of 0.00001. Adam optimizer is used with a batch size of 50. For model training using contrastive loss, the margin is set to 2. During triplet training, a margin of 0.4 is used. Code is implemented in PyTorch. All the experiments are performed on a DGX station with Intel Xeon CPU, 256 GB RAM, and four 32 GB Nvidia V100 GPU cards.

\begin{table*}[]
\centering
\renewcommand{\arraystretch}{1.2}
\caption{Identification accuracy (\%) of existing algorithms on the proposed IMFW dataset using the LightCNN-29 model.}
\label{Tab:ExistingLoss}
\begin{tabular}{|c|c|c|c||c|c|c||c|c|c|}
\hline
\multirow{2}{*}{\textbf{Algorithms}} & \multicolumn{3}{c||}{\textbf{Set 1 + Set 2}} & \multicolumn{3}{c||}{\textbf{Set 3}} & \multicolumn{3}{c|}{\textbf{Complete IMFW}} \\ \cline{2-10} 
 &
  \multicolumn{1}{l|}{\textbf{Rank 1}} &
  \multicolumn{1}{l|}{\textbf{Rank 5}} &
  \multicolumn{1}{l||}{\textbf{Rank 10}} &
  \multicolumn{1}{l|}{\textbf{Rank 1}} &
  \multicolumn{1}{l|}{\textbf{Rank 5}} &
  \multicolumn{1}{l||}{\textbf{Rank 10}} &
  \multicolumn{1}{l|}{\textbf{Rank 1}} &
  \multicolumn{1}{l|}{\textbf{Rank 5}} &
  \multicolumn{1}{l|}{\textbf{Rank 10}} \\ \hline 
\textbf{Pre-trained}                 & 52.77         & 73.37        & 87.96        & 66.17      & 91.17      & 92.64     & 53.97      & 71.02     & 76.70     \\ \hline
\textbf{Contrastive Loss}            & 62.96         & 83.33        & 92.59        & 77.94      & 92.64      & 95.58     & 63.06      & 80.68     & 85.22     \\ \hline
\textbf{Triplet Loss}                & 67.59         & 83.33        & 92.59        & 80.88      & 91.17      & 98.52     & 66.47      & 81.25     & 86.36     \\ \hline
\end{tabular}
\end{table*}

\subsection{Masked Face Recognition using Pre-trained Models}
For establishing the baseline performance on the IMFW dataset, four existing pre-trained deep face recognition models, VGGFace, ResNet50, LightCNN29, and ArcFace are used. LightCNN29 is pre-trained on the MS-Celeb-1M dataset \cite{guo2016ms}. The dataset contains approximately 10M images of 1M subjects. ArcFace is pre-trained on the CASIA-WebFace dataset \cite{yi2014learning}. The dataset contains 0.5M images of 10K subjects. Features extracted from the gallery and probe of the testing set are matched using Cosine distance. Table \ref{Tab:Pre-trained} shows the identification accuracy at rank 1, rank 5, and rank 10. For the experiments, we have combined the sets that are created by downloading images from the web. Therefore, results are reported on combined Set 1 and Set 2, Set 3, and the full dataset (IMFW). It is observed that existing deep models do not perform well for recognizing masked face images. ArcFace is one of the state-of-the-art face recognition models. However, it achieves only 62.96\%, 61.76\%, and 57.96\% identification accuracy at rank 1 on (Set 1 + Set 2), Set 3, and the full dataset, respectively. Fig. \ref{fig:CMC} shows the cumulative match characteristic (CMC) curves.  The low baseline performance indicates the need for sophisticated face recognition algorithms for masked face recognition in unconstrained settings.

\subsection{Masked Face Recognition using Existing Loss Functions}
This experiment is performed to evaluate the performance of existing loss functions on the proposed IMFW dataset. Two existing loss functions, contrastive loss and triplet loss, are used to train the LightCNN-29 model. Here, our aim is to match masked faces with their corresponding non-masked images in the gallery. Therefore, pairs are created by taking one masked image and one non-masked image during training using contrastive loss. Similarly, during model training using triplet loss, triplets are generated by taking non-masked images as the anchor and masked images as positive and negative. Fig. \ref{fig:Pipeline} shows the pipeline for masked face recognition. In the first step, facial regions are segmented using Tiny Face detector \cite{8099649} and resized to 128 $\times$ 128 resolution. Next, models are trained by minimizing existing loss functions to reduce the intra-class separation and increase the inter-class separability. During testing, features are extracted from the gallery and probe of the trained model. Finally, matching is performed using Cosine distance. Table \ref{Tab:ExistingLoss} shows the identification accuracy at ranks 1, 5, and 10 using the existing algorithms. It is observed that model training using existing algorithms enhance the performance at least by 10\% at rank 1. For instance, the identification accuracy at rank 1 on (Set 1 + Set 2) increases by 10.19\% and 14.82\% using contrastive loss and triplet loss, respectively, compared to the pre-trained model performance. It is important to observe that the performance of the algorithms is better on Set 3 compared to others. Set 3 contains images captured in both constrained and unconstrained settings, while the majority of the images in Sets 1 and 2 are taken in unconstrained settings. This further highlights the challenges of masked face recognition under unconstrained environmental conditions.

We have further benchmarked the proposed dataset against the LFWA dataset \cite{huang2008labeled}. For the experiment, we have followed the protocol similar to the proposed one. We have sampled 200 subjects with 1374 images from the LFWA dataset and divided the dataset into training and testing sets with non-overlapping subjects (70\% subjects in the training set and 30\% in the testing set). Using this protocol, the pre-trained LCNN29 model yields an accuracy of 96.11\% at rank 1 on the LFWA dataset. The fine-tuned LightCNN29 enhances the performance to 97.22\% rank 1 accuracy using triplet loss. However, on the IMFW dataset, LightCNN29 yields 53.97\% and 66.47\% rank 1 accuracies using pre-trained and triplet loss, respectively. The low performance on the IMFW dataset highlights the challenges of unconstrained masked face recognition in the Indian context.

\section{Conclusion}
COVID-19 pandemic has posed several challenges including masked face recognition. In the ``new normal’’, faces are covered at workplaces and public-places. In such a scenario, performing face recognition is a major challenge as existing algorithms generally do not perform well when the faces are obfuscated. The problem is further exacerbated when the attire (masks) diversity show very high intra-class and very low inter-class variations. This paper presents the Indian Masked Faces in the Wild database which includes the attire diversity and samples collected in unconstrained settings. Baseline experiments show that the masked face recognition is still an arduous task and require dedicated research efforts.

\section*{Acknowledgements}
This work is supported through research grants from RAKSHAK (iHub Drishti Foundation and DST) and MEITY. P. Majumdar is partly supported by DST Inspire Ph.D. Fellowship. M. Vatsa is also partially supported through Swarnajayanti Fellowship. The authors would like to express their gratitude towards Muskan Dosi for helping in the process of data collection.

% -------------------------------------------------------------------------
\bibliographystyle{IEEEbib}
\bibliography{Template}

\end{document}